\documentclass[10pt, a4paper]{article}
\usepackage{lrec-coling2024} 

\usepackage{natbib}
\usepackage{multibib}
\makeatletter
\def\@mb@citenamelist{cite,citep,citet,citealp,citealt,citepalias,citetalias}
\makeatother
\newcites{languageresource}{~}

\usepackage{graphicx}
\usepackage{tabularx}
\usepackage{soul}
\usepackage{latexsym}
\usepackage{CJKutf8}
\usepackage{float}
\usepackage{subfig}
\usepackage{overpic}
\usepackage{wrapfig}
\usepackage{inconsolata}
\usepackage{microtype}
\usepackage{inconsolata}
\usepackage{graphics}
\usepackage{graphicx}
\usepackage{sidecap}

\newcommand{\jam}{\begin{CJK*}{UTF8}{gbsn}堵车\end{CJK*}(traffic jam)}
\newcommand{\scenery}{\begin{CJK*}{UTF8}{gbsn}风景\end{CJK*}(scenery)}
\newcommand{\weather}{\begin{CJK*}{UTF8}{gbsn}天气\end{CJK*}(weather)}
\newcommand{\mood}{\begin{CJK*}{UTF8}{gbsn}心情\end{CJK*}(mood)}
\newcommand{\clothing}{\begin{CJK*}{UTF8}{gbsn}衣服\end{CJK*}(clothing)}
\newcommand{\bcn}{\begin{CJK*}{UTF8}{gbsn}}
\newcommand{\ecn}{\end{CJK*}}

\usepackage{titlesec}
\titleformat{\section}{\normalfont\large\bfseries\center}{\thesection.}{1em}{}
\titleformat{\subsection}{\normalfont\SmallTitleFont\bfseries\raggedright}{\thesubsection.}{1em}{}
\titleformat{\subsubsection}{\normalfont\normalsize\bfseries\raggedright}{\thesubsubsection.}{1em}{}
\renewcommand\thesection{\arabic{section}}
\renewcommand\thesubsection{\thesection.\arabic{subsection}}
\renewcommand\thesubsubsection{\thesubsection.\arabic{subsubsection}}

\usepackage{xcolor}
\usepackage{hyperref}
 \definecolor{darkblue}{rgb}{0, 0, 0.5}
  \hypersetup{colorlinks=true, citecolor=darkblue, linkcolor=darkblue, urlcolor=darkblue}

\usepackage{xstring}

\usepackage{color}

\title{Image Matters: A New Dataset and Empirical Study for Multimodal Hyperbole Detection}

\name{Huixuan Zhang, Xiaojun Wan} 

\address{Wangxuan Institute of Computer Technology, Peking University \\ State Key Laboratory of Media Convergence Production Technology and Systems \\
zhanghuixuan@stu.pku.edu.cn, wanxiaojun@pku.edu.cn}

\abstract{
Hyperbole, or exaggeration, is a common linguistic 
phenomenon. The detection of hyperbole is an important
part of understanding human expression. There have been several studies on hyperbole detection, but most of which focus on text modality only. However, with the development of social media, people can create hyperbolic expressions with various modalities, including text, images, videos, etc.
In this paper, we focus on multimodal hyperbole detection. We create a multimodal detection dataset from \emph{Weibo} (a Chinese social media) and carry out some studies on it. We treat the text and image from a piece of weibo as two modalities and explore the role of text and image for hyperbole detection. Different pre-trained multimodal encoders are also evaluated on this downstream task to show their performance. Besides, since this dataset is constructed from five different keywords, we also evaluate the cross-domain performance of different models. These studies can serve as a benchmark and point out the direction of further study on multimodal hyperbole detection.
 \\ \newline  }

\begin{document}

\maketitleabstract

\section{Introduction}
As defined by Merriam Webster, exaggeration means ``an act or instance of exaggerating something : overstatement of the truth'' \footnote{\url{https://www.merriam-webster.com/dictionary/exaggeration}} and hyperbole has the same meaning. Though hyperbolic expressions state something beyond fact or truth, they are not considered as lies. On the contrary, hyperbole is a common way of expressing one's strong feeling or opinion. As the second most used rhetorical device \citep{Kreuz1996}, the detection of hyperbole bears great importance for understanding human language and expression.

\begin{figure*}[htbp]
\centering
\subfloat[\bcn  这个天气是提前过冬  \ecn<emoji> <emoji>   \\ (Winter comes early.)]{
    \label{fig:1a}
    \includegraphics[width=0.2\textwidth]{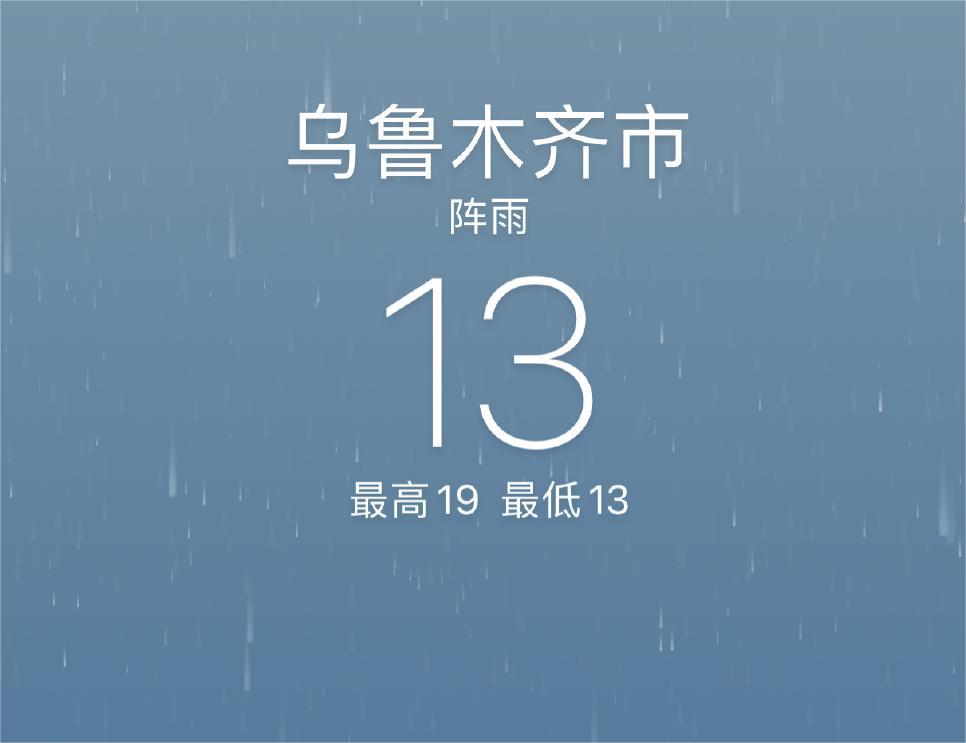}
}\hspace{5pt}
\subfloat[\bcn 天气太热了 \ecn \\ (It is too hot.)]{
    \label{fig:1b}
    \includegraphics[width=0.2\textwidth]{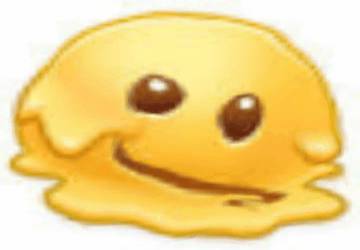}
}\hspace{5pt}
\subfloat[\bcn  心情是想杀了全世界  \ecn \\ (My mood is that I'd like to kill everyone in the world.)\\ (The words in the image says ``I did not piss off anyone.'')]{
    \label{fig:1c}
    \includegraphics[width=0.2\textwidth]{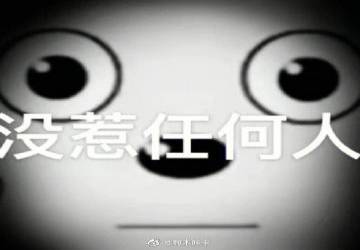}
}\hspace{5pt}
\subfloat[ \bcn 这个景区周六周日来就是要命！堵车能堵一小时 \ecn <emoji><url>  \\ (It's driving us to death to go to this scenic spot on weekend. We spent hours on traffic jams.) ]{
    \label{fig:1d}
    \includegraphics[width=0.2\textwidth]{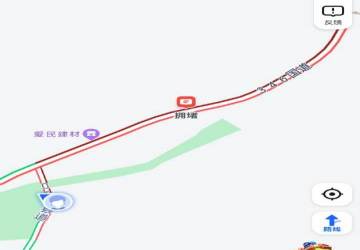}
}

\caption{Examples of image aiding the detection of hyperbole or serving as background information(all the examples are hyperbole). (a). In the image we can see that it is not so cold. (This is the surface of a weather application which says it is $13^\circ C$ in Urumqi. It's much colder in winter in Urumqi.) So we must refer to both text and image to correctly determine the post as hyperbole.; (b).The image is necessary because the text just states some fact. But with the melting face we can confidently label it as hyperbole because however hot it is, people do not melt.; (c). The grey background and stiff face in the image increases the confidence of annotating it as hyperbole.;(d).The image is just a screenshot of a navigating software. The hyperbolic meaning here comes from that people do not die from just going to this scenic spot, which cannot be revealed by this image. }
\label{fig:1}
\end{figure*}

Traditional studies focus more on text modality only. Their proposed datasets and methods considered mainly single short sentences (\citet{Troiano2018} \citet{Kong2020},\citet{Biddle2021})
However, the expression of hyperbole is not limited to text only. For example, we can hardly know whether the expression "\emph{Winter comes early.}" is hyperbolic or not unless we can see whether it is indeed so cold (Figure \ref{fig:1a}). Images, as widely seen on social media, serve an important role in expressing exaggerated meanings. They can serve as facts to reveal hyperbole contained in texts (Figure \ref{fig:1a}) as well as express hyperbole themselves (Figure \ref{fig:1b}) or together with texts (Figure \ref{fig:1c}). Meanwhile, some images can also just serve as background information and does no help to the expression of hyperbole (Figure \ref{fig:1d}), which adds additional difficulty to this task.

\emph{Weibo}\footnote{\url{https://weibo.com}} is one of the most popular social media in China and all over the world. With abundant publicly available posts with texts and images, it serves as the perfect resource for real-life multimodal hyperbole detection. We just need to automatically crawl enough posts from it and annotate them as hyperbole or not.

Our contributions in this paper can be summarized as follows: (1) We propose the first Chinese multimodal hyperbole detection dataset which can be used for further study.\footnote{Dataset can be found in \url{https://github.com/lleozhang/Multimodal_Hyperbole}.} (2) We analyze the dataset and get some statistical observations. We also summarize how two modalities (text and image) together express hyperbole. (3) We employ several neural methods on hyperbole detection, including unimodal and multimodal encoders, and reach these conclusions: (3.1) Image modality is more useful rather than misleading. With gating and attention mechanism, we can achieve relatively fine performance. We also point out that common sense is the potential direction to achieve better results.  (3.2) Compared with unimodal encoders and fusing methods, pre-trained multimodal methods perform badly on this downstream task. We analyzed its reason. (3.3) The dataset is built upon five different and disjoint keywords with generally disjoint contents. When applied on cross domains, the methods don't work very well. The systematic bias between posts with different keywords is the bottleneck hindering the generalization ability, especially for methods requiring deep interactions between modalities.

\section{Related Works}
\subsection{Studies on Hyperbole and Hyperbole Detection}
Not surprisingly, earlier studies on hyperbole focus more on its linguistic features. \citet{cano2009} made a detailed anylasis of the semantic features of hyperbole and first introduced a taxonomy which generally categorizes all exaggerations along two dimensions: quantity(overstating the  quantitaive or objective property of something, such as time, mass, etc.) and quality(overstating the qualitative or subjective property of something, such as feeling, emotion, etc.). These two types can be subcategorized into smaller ones, but that is beyond our discussion. 

Some multimodal analysis work has also been conducted. \citet{Ferré2009} manually analyzed how hyperbole is expressed with text, video and audio modalities. He focused on gestures, facial expressions, tones, etc. His work directly pointed out that other modalities can help reveal hyperbole contained in texts or words.

For the automatic detection of hyperbole, \citet{Troiano2018} created a hyperbole detection dataset HYPO, which contains 709 hyperbolic sentences and  their corresponding non-hyperbolic ones. They applied traditional machine learning methods and proposed some hand-crafted features (named QQ) to address this task.

\citet{Biddle2021} followed Troiano's work. They introduced triplet sampling and used triplet loss for forcing their model to differentiate hyperbolic and non-hyperbolic sentences. They used BERT as the backbone text encoder instead of pre-trained word vectors(used by Troiano's work). 

As for Chinese hyperbole detection, \citet{Kong2020} followed Troiano's idea and created the first Chinese hyperbole detection dataset HYPO-cn. They analyzed statistical features of the sentences in the dataset thoroughly. They also manually analyzed strategies humans used to express hyperbole. They compared several traditional and deep-learning methods on this dataset. They reached the conclusion that deep-learning methods outperform traditional methods on this task.

However, to the best of our knowledge, there hasn't been any study on multimodal hyperbole detection.

\subsection{Other Relevant Multimodal Tasks}
Multimodal sarcasm detection has become a hot topic in recent years. \citet{Cai2019} first proposed a dataset built on Twitter, which is still widely used by researchers. They also proposed a hierarchical fusion model to address this task. \citet{xu2020} followed this work and proposed D\&R Net. \citet{Liang2021} introduced in-modal and cross-modal graphs(InCrossMGs) to leverage the ironic expressions from both in- and cross-modal perspectives by extracting the dependency relations of fine-grained phrases or patches within the distinct modality and across multiple modalities. \citet{Liu2022} leverages graph attention networks and image captions as knowledge enhancement. 

On the other side, \citet{Santiago2019} focused on video and text modalities and proposed a sarcasm detection dataset named MUStARD. MUStARD is compiled from popular TV shows and consists of audiovisual utterances annotated with sarcasm labels along with its corresponding context. \citet{Wu2021} continued on this work and proposed IWAN to model the word-level incongruity between modalities via a scoring mechanism.

Multimodal sentiment analysis is also a widely-discussed topic. \citet{Wang2020} proposed an end-to-end fusion method based on transformers. \citet{Yang2021} introduced multi-channel graph neural networks. \citet{Xue2022}, \citet{Du2022} focused on attention-based methods such as gated attention. Inspirations can be gained from these methods.

Multimodal hyperbole detection bears both similarities and differences with the tasks mentioned above. It requires the understanding and interaction of both modalities. However, the relation between the two modalities can be rather complicated. This will be discussed in Section \ref{sec:4}.

\section{Dataset Creation}

\subsection{Hyperbole Definition}
To avoid ambiguity, we need to give a clear definition to hyperbole. Based on the definition given by Merriam Webster \footnote{\url{https://www.merriam-webster.com/dictionary/exaggeration}} and \citet{Troiano2018}, we here define hyperbole as ``an expression (words, images, etc) that goes significantly beyond fact or common sense but not taken as lies''. A post is considered hyperbolic if and only if it contains hyperbolic expression.

For example, the post ``\emph{I'm dying of heat}'' is hyperbolic because normally most people do not easily die of heat. But if there is a post ``\emph{5 workers died of extreme heat}'' with real pictures, the post is certainly not considered hyperbolic. However, if a post says ``\emph{100 billion workers died of extreme heat}'', it is then considered neither hyperbole or non-hyperbole. Instead, it is probably a lie. As for images, Figure \ref{fig:1b} provides a perfect example. Since people do not melt however hot it is, a melting face representing the unbearably hot weather is typical hyperbole. 

\subsection{Data Collection and Preprocessing}
\label{sec:32}
\emph{Weibo} is a popular Chinese social medium. We automatically crawl about 10000 posts from \emph{Weibo} with 5 keywords: \emph{\jam}, \emph{\scenery}, \emph{\weather}, \emph{\mood}, \emph{\clothing}.
Each keyword contains roughly same number of posts (about 2000). We choose the five keywords because they are generally disjoint and close to daily lives. We carefully set criteria so that each post crawled is original and with at least one image. 

After collecting these posts, we first clean up the data. The urls in the texts are replaced with a specific token \emph{<url>}. Some emojis are actually hyperlinks to images, so they are replaced with another \emph{<emoji>} token. Hashtags (words or phrases led by \emph{\#}) are not removed because there is no obvious distinction between hashtags in hyperbolic and non-hyperbolic posts according to our observation. Some posts have more than one image, so we randomly select one image and drop the others. Since this is done before any human work, there is no concern that it will introduce any inconsistency or bias. Then we conduct a rough selection. Posts with too long or unreadable texts or unclear pictures are manually dropped to save some annotation work. (Here we define “too long” as posts longer than 200 characters.) We set the limitation because longer posts are harder for annotation. Annotators may miss the hyperbolic part of a too long post thus leading to unreliable results. We believe this setting makes our dataset cleaner. Advertisement posts that appear repeatedly are also dropped to avoid polluting the dataset. 

\subsection{Data Annotation}
We employ several human annotators to annotate these posts as hyperbole or non-hyperbole. All of them are Chinese college students. They are required to follow an annotation specification given by us. We first ask them to label 200 randomly chosen posts, and select two of them whose annotation results fit the specification best. We communicate with them about their mistakes to help them understand our definition of hyperbole better. They then annotate the rest posts. We manually check each inconsistent case and discuss with them to determine whether each is hyperbole. 

Since there are much more non-hyperbolic posts, we randomly choose from them and create the final dataset in which \#hyperbole:\#non-hyperbole is approximately 50\%:50\%. Some basic statistics of the final dataset are listed in Table \ref{tab:2}. 

\begin{table}
\centering
\begin{tabular}{ccc}
\hline
\textbf{Keyword} & \textbf{Total Count} & \textbf{Hyperbole}\\
\hline
\jam  & 496 & 252 \\
\scenery & 402 & 204 \\
\weather & 396 & 194 \\ 
\mood & 451 & 207 \\
\clothing & 442 & 198 \\
total & 2160 & 1055 \\\hline
\end{tabular}
\caption{Basic statistics of our dataset}
\label{tab:2}
\end{table}

\section{Dataset Analysis}
\label{sec:4}
In this section, we make some analysis on our constructed dataset. 

First, we consider the length of the text. In Chinese, the length is simply the number of characters in the text. We assume that hyperbolic posts and non-hyperbolic posts may have similar text length. But the result turns out that hyperbolic posts have significantly longer texts than non-hyperbolic ones. What's more, compared with non-hyperbolic ones, hyperbolic posts of different keywords also have more varied text length. The average text length of different keywords are shown in Table \ref{tab:3}. As can be seen, non-hyperbolic posts have texts with relatively similar lengths among different keywords, while hyperbolic posts have quite varied text length among keywords.  

\begin{table*}
\centering
\small
\begin{tabular}{ccccccc}
\hline
\textbf{} & \textbf{\jam} & \textbf{\scenery} & \textbf{\weather} & \textbf{\mood } & \textbf{\clothing}  \\
\hline
non-hyperbole & 61.3 & 56.8 & 57.7 & 64.3 & 70.9 \\ 
hyperbole & 81.0 & 96.1 & 79.3 & 65.5 & 88.3 \\
\hline
\end{tabular}
\caption{Average text lengths of different keywords. }
\label{tab:3}
\end{table*}

\begin{figure}[htbp]
    \centering
    \small
    \begin{minipage}[l]{0.2\textwidth}
        \includegraphics[width=1.0\textwidth]{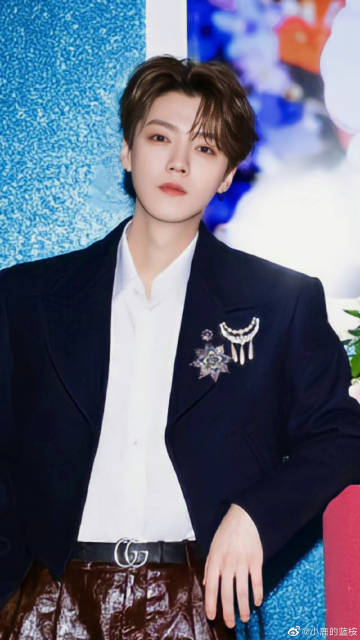}
    \end{minipage}%
    \hspace{5pt}
    \begin{minipage}[r]{0.2\textwidth}
           \bcn <url> <url>鹿晗，晚安<url>\#全世界最帅的老板\# 鹿晗<url> \#gucci品牌代言人鹿晗\# <url> @M鹿M  你是我看不完的风景，唱不完的情歌，写不出的浪漫，永远都想捧在手心里的宝藏男孩  \ecn  \\ (Good Night, Lu Han. \#The most handsome boss in the world\# Lu Han. \#Gucci brand spokesman Lu Han.\# You are my endless scenery, endless love song, endless romance, the treasure boy I want to hold in my hands forever.)
    \end{minipage}
    
    \caption{An example of hyperbolic post expressing their love towards a star with the word \emph{\scenery}.(There are many much longer examples, for conciseness we choose a relatively short one here.)}
    \label{fig:10}
\end{figure}

We guess that there are two main reasons leading to this phenomenon. First, hyperbolic expressions are usually related with strong emotions. People tend to use more words to fully express their strong feelings. The other reason is that some hyperbolic sentences with certain keywords bear some kind of similarity. Taking figure \ref{fig:10} as an example, there are several hyperbolic posts of keyword \emph{\scenery} with this pattern to express their love towards a star. This pattern does not appear in other keywords or in most non-hyperbolic posts, which partly explains why hyperbolic posts contain longer texts and why hyperbolic posts have varied text length among different keywords. 

Then we focus on lexical features of hyperbolic posts. There's no doubt that some words appear more often in hyperbolic posts than non-hyperbolic ones. To quantitatively examine this phenomenon, We calculate smoothed chi-score \citep{Yang1997} $\tilde{K}(d)$ of each word $d$. The bigger $\tilde{K}(d)$ is, the higher correlation between $d$ and hyperbole is. For details about smoothed chi-score, please refer to appendix \ref{sec:app-1}. Since there are no spaces between words in Chinese, we use \emph{Spacy} \footnote{\url{https://github.com/explosion/spaCy}} to separate words and remove stop words. We calculate $\tilde{K}(d)$ for all remaining words. The top-10 words are listed in Table \ref{tab:4}. As can be seen, one can easily think of hyperbolic expressions with some words such as  \bcn 死 \ecn  (death). Other words are corresponding to certain keywords, such as  \bcn 冷 \ecn  (cold) is closely related to \bcn  天气 \ecn  (weather). However, when observing these words alone, most of them does not contain hyperbolic meaning themselves, which means that there is no reason to determine whether a post is hyperbole through some kind of `` keywords ``, thus further proving the difficulty of this task. It is also interesting that though we do not consider emotional exaggeration(such as the expression ``Ah Ah Ah it's so cold!'') as real hyperbole (According to our definition, this expression says nothing beyond fact),  \bcn 啊啊 \ecn  (AhAh) is still closely related to hyperbole. It meets our assumption above that hyperbole is related to strong emotions. 

 \begin{table}[]
     \centering
     \small
     \begin{tabular}{cc}
        \hline
        word($d$) & $\tilde{K}$(d)\\
        \hline
         \bcn 死 \ecn  (death)  &  0.00412 \\
         \bcn 冷 \ecn  (cold)  &  0.00372 \\
         \bcn 买到 \ecn  (bought)  &  0.00372 \\
         \bcn 腿 \ecn  (leg)  &  0.00359 \\
         \bcn 弄 \ecn  (cause)  &  0.00359 \\
          \bcn 适合 \ecn  (fit)  &  0.00342 \\
         \bcn 啊啊 \ecn  (AhAh)  &  0.00342 \\
         \bcn 加 \ecn  (plus)  & 0.00342 \\
         \bcn 头发 \ecn  (hair)  &  0.00342 \\
         \bcn 开学 \ecn  (term begins)  &  0.00342 \\
        \hline
     \end{tabular}
     \caption{Top 10 words with highest $\tilde{K}(d)$}
     \label{tab:4}
 \end{table}

\begin{figure}[htbp]
    \centering
    \small
    \begin{minipage}[l]{0.2\textwidth}
        \includegraphics[width=1.0\textwidth]{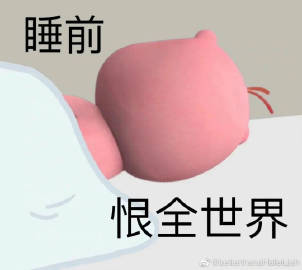}
    \end{minipage}%
    \hspace{5pt}
    \begin{minipage}[r]{0.2\textwidth}
         \bcn 完全没心情在社交平台活跃。物流走了2个半月了还没到，我的心好疲惫。 \ecn \\ (Completely no mood for staying active on social media. It's been two and a half months and my express still hasn't arrived. My heart feels so tired.)
    \end{minipage}
    
    \caption{An example of images with hyperbolic words in it. (The words in the images say ``I hate the whole world before sleep.``)}
    \label{fig:5}
\end{figure}

Finally, we consider the roles that images play in expressing hyperbole. We ask the annotators to check all the images alone, trying to determine whether the post can be classified as hyperbole with image modality only. We find out that in all hyperbolic posts, only 15.8\% images are themselves hyperbolic (like the one in Figure \ref{fig:1b}). We also need to note that some images are considered hyperbolic because they contain hyperbolic texts (Figure \ref{fig:5}). We count roughly and find there are only 4\% of all hyperbolic posts that belong to this category, which is acceptable. This result reveals that images mostly serve as an assistant when expressing hyperbole rather than being hyperbolic themselves. As is discussed above, images can serve as ground truth to reveal hyperbole (Figure \ref{fig:1a}) or give us higher confidence to determine one post as hyperbole (Figure \ref{fig:1c}). According to this observation, we may question whether introducing image modality is helpful in recognizing hyperbole, which will be discussed in detail in \ref{sec:51}.

In summary, image and text can express hyperbole in the following ways: (1) one modality expresses hyperbole itself and the other modality serves as background information (\emph{case-I});  (2) two modalities together express hyperbolic meanings, one expressing hyperbole and the other serving as ground truth to reveal this hyperbole (\emph{case-II}); (3) two modalities together express hyperbolic meaning, one expressing hyperbole and the other offering extra confidence (\emph{case-III}).

\section{Empirical Studies on Hyperbole Detection}
We treat the task of multimodal hyperbole detection as a binary classification task with hyperbole as positive sample. In this section, we aim to investigate and answer the following three questions related to this task:
\begin{itemize}
    \item \textbf{RQ1:} Is image modality more useful rather than misleading in multimodal hyperbole detection? 
    \item \textbf{RQ2:} Are pre-trained multimodal encoders effective on this task?
    \item \textbf{RQ3:} Can typical methods achieve good performance on cross-domain circumstances?
\end{itemize}

In the next three subsections, we will conduct empirical studies to discuss the above three questions, respectively.

\subsection{Experiments on Different Modalities (RQ1)}

As discussed in Section 4, images mostly serve as assistants rather than express hyperbole themselves. So it is natural to ask: is image modality more useful than misleading in hyperbole detection? If the answer is yes, which way is more suitable for utilizing image? For example, is deep modality fusion more helpful than shallow ones?

We describe our methods from shallower fusion to deeper fusion. First, we use two unimodal encoders: BERT \citep{Delvin2019} and ResNet50 \citep{He2016}, to encode text and image modality individually. Specifically, we use BERT that is pre-trained on Chinese\citep{Cui2020}. We fine-tune them to achieve better performance.  

After unimodal encoding, we apply different fusion methods on this task. Denote $f_{text}$ as the feature vector of text (outputted by BERT) and $f_{image}$ as the feature vector of image (outputted by ResNet), the most trivial idea is simply concatenating $f_{text}$ and $f_{image}$ (We name this method \emph{concat}). A slight improvement on this idea is implementing a gating method to select more significant features. We name this method \emph{gate}. To be more detailed, consider two feature vectors $f_{image}$ and $f_{text}$, the gating method we use can be summarized as:
\[
    g_{text}=\sigma(W_{text}f_{image}+b_{text})
\]
\[
    g_{image}=\sigma(W_{image}f_{text}+b_{image})
\]
\[
    output=concatenate(g_{text} \otimes f_{text}, g_{image} \otimes f_{image})
\]
where $W_{text}, W_{image}, b_{text}, b_{image}$ are learnable parameters, $\sigma$ is the \emph{Sigmoid} function ($\sigma(x)=\frac{1}{1+e^{-x}}$). 

\begin{figure*}[htbp]
\centering
\includegraphics[width=1.0\textwidth]{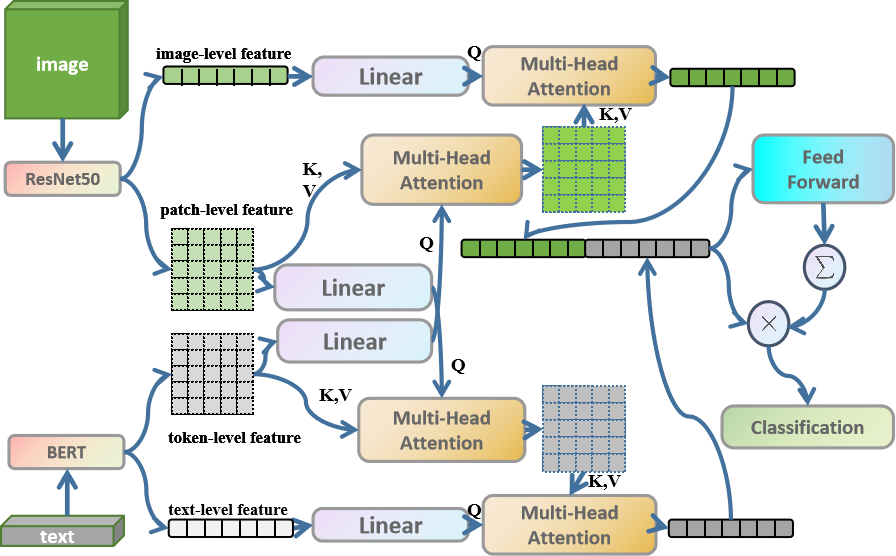}
\caption{The architecture of the \emph{attn-gate} fusion model.}
\label{fig:6}
\end{figure*}

Attention is a widely used method which can utilize fine-grained features. We utilize both text-level and token-level features outputted by BERT as well as both image-level and patch-level features outputted by ResNet. We implement a multi-head attention mechanism to extract cross-modality fine-grained features. To be more detailed, we use the fine-grained features (i.e., token-level or patch-level features) of one modality as $Q$, the fine-grained features of the other modality as $K$ and $V$. Thus we obtain two cross-modality fine-grained feature representations. We use another two multi-head attention modules to further aggregate these fine-grained feature representations respectively (using image- or text-level feature representation as $Q$ and the corresponding cross-modality fine-grained feature representation as $K$ and $V$) and get the encoding vector of either modality. We then concatenate these two encodings and implement a gating mechanism similarly (We name this method \emph{attn-gate}.) The main architecture of this method is shown in Figure \ref{fig:6}. 

\label{sec:51}

After fusing with either of the above methods, we feed the feature vector into several fully connected layers to finish the final classification. We use cross-entropy as loss function and Adam \citep{Kingma2014} as optimizer.  All these models share the same hyper-parameters. We believe the above methods are enough to answer our question on this relatively small dataset.

We report 10-fold cross validation results of the models to ensure the reliability of our results. Our hyper-parameters are listed in Table \ref{tab:hyp-para}.The average results are shown in Table \ref{tab:5}.

\begin{table}[htbp]
\centering
    \begin{tabular}{c|c}
    \hline
        Batchsize & 32 \\
        Learning rate & 1e-4 \\
        Epochs & 100 \\
        Dropout & 0.5 \\
        Random seed & 1008 \\
        Number of heads & 4 \\
    \hline
    \end{tabular}
    \caption{Hyper parameters used.}
    \label{tab:hyp-para}
\end{table}

\begin{table}[htbp]
    \centering
    \begin{tabular}{c|cc}
    \hline
        \textbf{model} & \textbf{F1-score} & \textbf{Accuracy} 
        \\
    \hline
        BERT  & $0.702_{\pm 0.174}$ & $0.713_{\pm 0.027}$ \\ 
        ResNet & $0.553_{\pm 0.078}$ & $0.495_{\pm 0.028}$  \\
        Concat & $0.734_{\pm 0.199}$ & $0.743_{\pm 0.017}$  \\
        Gate & $0.742_{\pm 0.193}$ & $0.749_{\pm 0.029}$ \\
        \textbf{Attn-Gate} & $\textbf{0.750}_{\pm 0.178}$ & $\textbf{0.758}_{\pm 0.018}$  \\
    \hline
    \end{tabular}
    \caption{Average results with 10-fold cross-validation (BERT and ResNet use only single-modality information.). The subscripts show the standard deviation of each experiment.}
    \label{tab:5}
\end{table}

Some conclusions can be drawn from the results. First, with single modality, text-only model (BERT) performs much better than image-only model (ResNet). This is compatible with our analysis that most images are not hyperbolic themselves.  Therefore although we've already discussed the importance of image modality for humans to determine hyperbole, it's still questionable that whether image modality is more useful rather than misleading under this circumstance. 

Our following result proves that image modality is indeed more useful rather than misleading.  Though with text modality only can achieve a fine result, introducing image modality, even simply giving it the same significance as text modality (\emph{concat}), still apparently increases the performance. Comparing with \emph{BERT}, this improvement is \emph{statistically significant}. This result clearly shows that introducing image modality together with text modality is useful for determining hyperbole.

Further, gating mechanism may be useful for increasing model performance. This insight can be drawn from our analysis of the dataset. Since under most circumstances, there are probably some features serving as background information, we should give these features less focus. As can be seen from the results, even with trivial gating mechanism, the result is slightly improved compared with simply concatenating the two modalities' outputs. However, this trivial gating method still has its own problems. First, using BERT only can achieve relatively fine results, which shows that text modality contains more information for detecting hyperbole. Thus we observe that the trivial gating method tends to give text modality a higher priority. This systematic bias leads to insufficient utility of image modality. 

\begin{figure}
    \centering
    \subfloat[ \bcn 这个天气是提前过冬 \ecn  \\(Winter comes early.)]{
        \centering
        \includegraphics[width=0.2\textwidth]{1a.jpg}
        \label{fig:7a}
    }\hspace{5pt}
    \subfloat[ \bcn 想看个天气参考一下买什么衣服。看完后，算了，不买了， 这天还穿什么衣服 <emoji>  \ecn \\(Referring to weather forecast to see which clothes to buy. Now I think there's no need wearing any clothes in such weather. )]{
        \centering
        \includegraphics[width=0.2\textwidth]{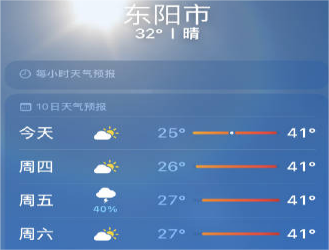}
        \label{fig:7b}
    }
    \caption{Two examples of hyperbole cases.}
    
\end{figure}

Another problem is that \emph{gate}, as well as \emph{concat}, is unable to model the fine-grained relationship between text and image. Figure \ref{fig:7a} gives an example. All other methods except \emph{attn-gate} failed on this case. Since they just simply use high-level features of two modalities, they are unable to notice the connection between certain tokens(\emph{Winter}) and a certain feature(the temperature in the image).  In contrast, \emph{attn-gate} is able to extract fine-grained features between two modalities. In fact, comparing with the trivial \emph{concat} method, \emph{attn-gate} is \emph{statistically significantly} better. Since we performed 10-fold cross validation, we believe that this improvement clearly shows that even though most hyperbole are not directly expresses through images, introducing images, especially using deep fusion, is still helpful in recognizing hyperbole. 

There are also cases where all these methods failed. Figure \ref{fig:7b} is a typical failure case. The reason is that these models do not know even under such hot weather, people still need to wear clothes. The expression ``there's no need wearing any clothes'' is an overstatement of how hot it is. Since models do not have any common sense knowledge, they are unable to recognize this kind of hyperbole, which leads to the failure.

In summary, our answer to this question (RQ1) is that: images are more useful rather than misleading. With gating and attention mechanism, we can more accurately utilize fine-grained fused features between two modalities, which can help identify hyperboles of \emph{case-II} and \emph{case-III}. However, without common sense, the model can never fully understand hyperbole. To further improve the performance, the key is introducing common sense into the models. 

\subsection{Pre-trained Multimodal Encoders Evaluation (RQ2)}
\label{sec:52}
Pre-trained multimodal models are proved effective in many downstream tasks including zero-shot learning, image-text retrieval, VQA, etc. However, when speaking of multimodal classification tasks, few discussions are made before. In this subsection, we apply two pre-trained models CLIP \citep{Radford2021} and BriVL \citep{Huo2021} on this downstream task to evaluate their performance. CLIP \citep{Radford2021} is a popular multimodal encoder and BriVL \citep{Huo2021} is a similar one that is pre-trained on Chinese. Besides, BriVL claims to use a more advanced algorithm(inspired by MoCo \citep{He2020}) to obtain better encodings of both modalities. 

We still report 10-fold cross-validation results. Specifically, we use CLIP that is pre-trained on Chinese texts \citep{Yang2022}. For BriVL, we directly use the open-source code and tools\footnote{\url{https://github.com/BAAI-WuDao/BriVL}}.

We try two ways to use these pre-trained models. One is giving prompts (\emph{prompt}) to the text (We add prompts `` \bcn 这是夸张 \ecn  ''(This is hyperbole.) and `` \bcn 这不是夸张 \ecn  ''(This is not hyperbole) at the beginning of each text. So we get two texts with different prompts and same content.). To be more detailed, for each case we encode one image and the two texts with different prompts, and try to maximize the cosine similarity between the image and the text with correct prompt. This is a common method used in image classification tasks.

\begin{figure*}[htbp]
\centering
\subfloat[The main architecture of the prompt method.]{
    \label{fig:9a}
    \includegraphics[width=0.4\textwidth]{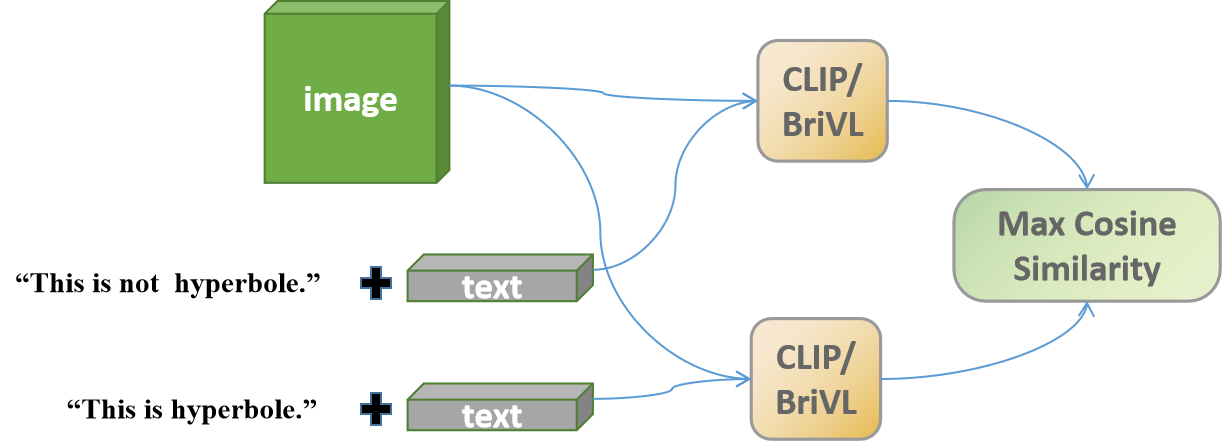}
}\hspace{5pt}
\subfloat[The main architecture of pre-trained models used as encoders.]{
    \label{fig:9b}
    \includegraphics[width=0.4\textwidth]{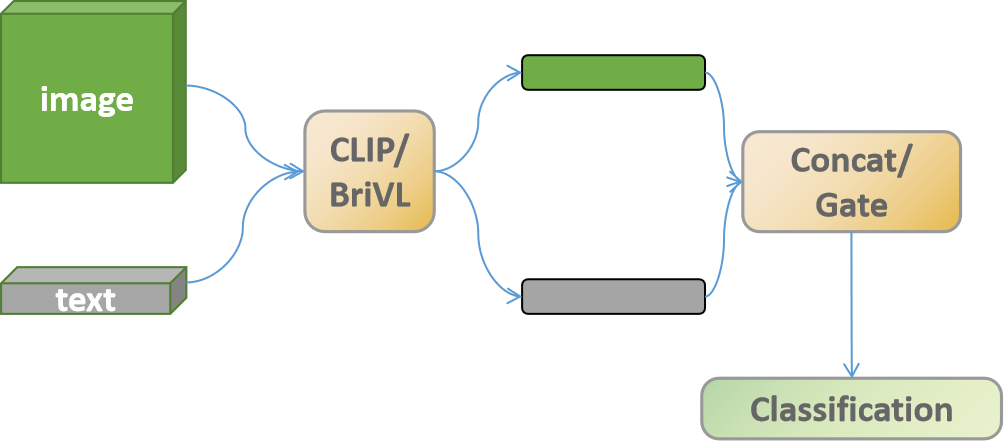}
}\hspace{5pt}

\caption{The architecture of the two methods using pre-trained models.}
\label{fig:9}
\end{figure*}

Another way is using these models simply as encoders and treat their output as feature vectors. Then we concatenate them (\emph{concat}) or use gating methods (\emph{gate}) just like what we have done above. In this way no prompts are added to texts. Both the two ways are seldom discussed in similar tasks like multimodal sarcasm detection previously. Details of both ways are shown in figure \ref{fig:9}.

We fine-tune theses models to achieve better results. However, the result turns out to be very disappointing. As can be seen in Table \ref{tab:6}, they perform much worse than most methods mentioned in Section \ref{sec:51}. 

\begin{table}[htbp]
    \centering
    \begin{tabular}{c|cc}
    \hline
        \textbf{model} &  \textbf{F1-score} & \textbf{Accuracy} \\
        \hline
        CLIP+prompt & $0.632_{\pm 0.159}$ & $0.642_{\pm 0.042}$\\
        CLIP+concat & $0.584_{\pm 0.165}$ & $0.644_{\pm 0.035}$\\
        CLIP+gate & $0.580_{\pm 0.174}$ & $0.642_{\pm 0.025}$\\
        BriVL+prompt & $0.587_{\pm 0.170}$ & $0.628_{\pm 0.025}$ \\
        BriVL+concat & $0.665_{\pm 0.165}$ & $0.667_{\pm 0.038}$\\
        BriVL+gate & $0.644_{\pm 0.162}$ & $0.637_{\pm 0.031}$ \\
        \hline
    \end{tabular}
    \caption{Results of multimodal models.}
    \label{tab:6}
\end{table}

We analyze why the results are much worse. \citet{Radford2021} has already pointed out that CLIP performs badly on classifying abstract concepts like determining whether a scene is normal or not. Hyperbole detection is a typical task involving abstract concept, so it is not surprising that \emph{prompt} does not work well on this task. What's more, since this kind of task involves classifying image and text together rather than image itself, the prompt method is even more unsuitable. We believe the same reason works for BriVL. 

Besides, the text in this kind of task is usually not a simple description of the image. It is a common case that there is no strong semantic correlation between corresponding text and image, which is quite different from the dataset CLIP is trained on. As is discussed in \citet{Radford2021} and \citet{Huo2021}, CLIP performs badly under this circumstance. \citet{Huo2021} claims to have tried to avoid assuming this strong correlation in BriVL. As can be seen, it turns out to be effective to some extent when compared with CLIP. However, some texts in this task have very little or even no explicit semantic correlation with corresponding images. We believe that this bias probably leads to poor encoding of either modality and causes the bad performance.

\subsection{Cross-domain Experiments (RQ3)}

In this part, we briefly evaluate the methods mentioned in Section \ref{sec:51} on cross-domain hyperbole detection. As is shown in Section \ref{sec:32}, the posts are of five different keywords and their contents are generally disjoint. We use posts from three keywords for training, one for validating and one for testing. For each method, we repeat this procedure five times to test on all five keywords and report average performance. The results are shown in Table \ref{tab:7}. Since pre-trained multimodal models perform badly, we do not discuss them in this part.

\begin{table}[htbp]
    \centering
    \begin{tabular}{c|cc}
    \hline
        \textbf{model} & \textbf{F1-score} & \textbf{Accuracy} \\
        \hline
        BERT & $0.652_{\pm 0.162}$ & $0.667_{\pm 0.045}$ \\
        ResNet & $0.513_{\pm 0.079}$ & $0.500_{\pm 0.020}$ \\
        Concat & $0.700_{\pm 0.168}$ & $0.711_{\pm 0.028}$ \\
        Gate & $0.683_{\pm 0.161}$ & $0.704_{\pm 0.037}$ \\
        Attn-Gate & $0.700_{\pm 0.167}$ & $0.714_{\pm 0.043}$\\
        \hline
    \end{tabular}
    \caption{Results of cross-domain experiments. The subscripts show the standard deviation of each experiment.}
    \label{tab:7}
\end{table}

\begin{figure}
    \centering
    \subfloat[ \bcn 骑上我心爱的小自行车，它不会堵车，但会被罚款 \ecn  \\(Riding on my lovely little bike. It won't be stuck in a traffic jam, but I may be penalized.)]{
        \centering
        \includegraphics[width=0.2\textwidth]{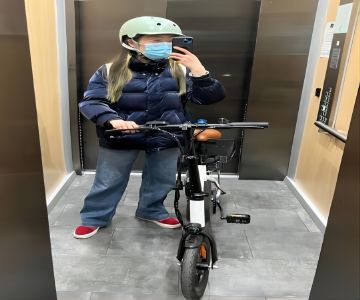}
        \label{fig:8a}
    }\hspace{5pt}
    \subfloat[ \bcn 让我晚回家半个小时的堵车 <emoji> \ecn  \\(The traffic jam that makes me arrive half an hour later than usual.)]{
        \centering
        \includegraphics[width=0.2\textwidth]{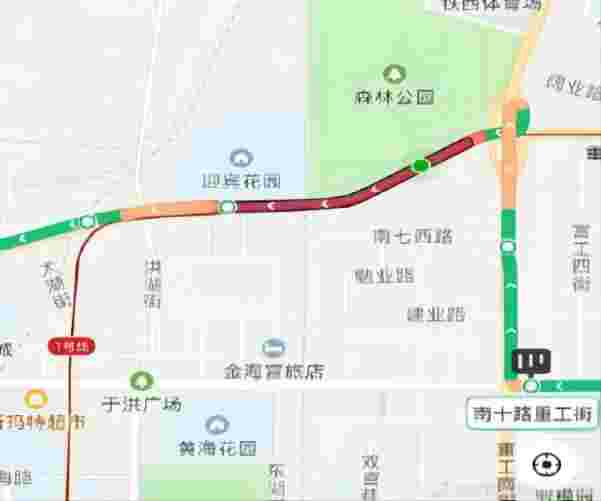}
        \label{fig:8b}
    }
    \caption{Two examples of test cases (Both of them are not hyperbole).}
    \label{fig:8}
\end{figure}

As can be seen, all methods perform worse than in Section \ref{sec:51}, which meets our expectation because cross-domain hyperbole detection is certainly a tougher task. The results are not that terrible, which shows the generalization ability of the methods proposed above. Introducing image modality still apparently increases the performance as can be seen in \emph{concat}. However, we are surprised to find that gating and attention mechanism do not bring higher performance. We take two examples (Figure \ref{fig:8}) from the keyword  \jam  (traffic jam) to illustrate the reason. 

As can be seen in Figure \ref{fig:8a}, the sentence contains an expression \emph{won't be stuck in a traffic jam} which is seemingly too absolute here. However, in this case, it is probably a fact that bikes will not be stuck in traffic jams, so it is not hyperbole. When trained on other keywords, the models will not have such knowledge (``bikes don't get stuck in traffic jams``), so they naturally fail on this case. This explains why all methods perform worse on this task. Similarly, the screenshot of a navigating software only appears in posts with the certain keyword \jam, so the models have no idea of what it is. Therefore, models requiring deep interaction between text modality and image modality may be confused more seriously. We believe that this systematic inconsistency is the reason why deep fusion methods perform much worse on this task. And again, common sense knowledge is still helpful in easing this problem.

\section{Conclusion and Future Work}
In this paper, we first introduce a new task of multimodal hyperbole detection and build a dataset for our study. We point out that different modalities can express hyperbole in mainly three ways. We conduct several experiments and prove that image modality is useful in hyperbole detection. We point out the importance of deep fusion of two modalities. Pre-trained multimodal models are proved ineffective on this task. Cross-domain results show that the methods still work relatively fine while the inconsistency between posts with different keywords is certainly the bottleneck of this task. For future work, we plan to make use of common sense knowledge in our model to achieve a better result. 

\section{Ethics Statement}
There is not any discrimination in this work. All the examples shown in this paper are just for academic use and do not represent our attitude. We respect different human races and different aesthetic judgments. However, the posts contained in our dataset are automatically crawled from the Internet, so there may be potential controversial statements and dirty words. We sincerely apologize to those who may feel offended. 

\section*{Acknowledgement}
 This work was supported by Beijing Science and Technology Program (Z231100007423011), National Key R\&D Program of China (2021YFF0901502), National Science Foundation of China (No. 62161160339) and Key Laboratory of Science, Technology and Standard in Press Industry (Key Laboratory of Intelligent Press Media Technology). We appreciate the anonymous reviewers for their helpful comments. Xiaojun Wan is the corresponding author.
 
\section{Bibliographical References}\label{reference}

\bibliographystyle{lrec-coling2024-natbib}
\bibliography{main}

\appendix
\section{Smoothed Chi-Score}
\label{sec:app-1}
Consider a certain word $d$, the chi-score of the word $K(d)$ is 
\[
     K(d)=\frac{(A+B+C+D)(AD-BC)}{(A+C)(B+D)(A+B)(C+D)}
\], where $A$ is the number of hyperbolic posts that contain $d$, $B$ is the number of non-hyperbolic posts that contain $d$, $C$ is the number of hyperbolic posts that do not contain $d$ and $D$ is the number of non-hyperbolic posts that do not contain $d$. To smooth it, we use 
\[
    \tilde{A}=A+1,\tilde{B}=B+1,\tilde{C}=C+1,\tilde{D}=D+1  
\]. So we have
\[
    \tilde{K}(d)=\frac{(\tilde{A}+\tilde{B}+\tilde{C}+\tilde{D})(\tilde{A}\tilde{D}-\tilde{B}\tilde{C})}{(\tilde{A}+\tilde{C})(\tilde{B}+\tilde{D})(\tilde{A}+\tilde{B})(\tilde{C}+\tilde{D})}
\].
\end{document}